\documentclass[11pt]{article} %

\usepackage[utf8]{inputenc} %
\usepackage[T1]{fontenc}    %
\usepackage{hyperref}       %
\usepackage{url}            %
\usepackage{booktabs}       %
\usepackage{amsfonts}       %
\usepackage{nicefrac}       %
\usepackage{microtype}      %
\usepackage{xcolor}

\usepackage[margin=1in]{geometry}

\usepackage[round,semicolon]{natbib}

\usepackage[utf8]{inputenc} %
\usepackage[T1]{fontenc}    %
\usepackage{hyperref}       %
\hypersetup{
    colorlinks,
    breaklinks,
    linkcolor = blue,
    citecolor = blue,
    urlcolor  = blue,
}
\usepackage{url}            %
\usepackage{amsfonts}       %
\usepackage{nicefrac}       %
\usepackage{microtype}      %
\usepackage{amsmath,bm}
\usepackage{amsthm}
\usepackage[algo2e, ruled, vlined]{algorithm2e}
\usepackage{algorithm}
\usepackage{bbm}      
\usepackage{mathtools}  \usepackage{matlab-prettifier}
\usepackage{amsopn}
\usepackage{amssymb}
\usepackage{graphicx}
\usepackage{xcolor}
\usepackage{footnote}
\usepackage{makecell}
\makesavenoteenv{tabular}
\makesavenoteenv{table}

\usepackage{booktabs,caption,threeparttable}

\usepackage{xcolor}
\usepackage{soul}
\usepackage{changepage}

\definecolor{Green}{rgb}{0.13, 0.65, 0.3}
\definecolor{Amber}{rgb}{0.3, 0.5, 1.0}
\usepackage[]{color-edits}
\usepackage{comment}

\usepackage{lipsum}
\usepackage{minitoc}
\usepackage{subcaption}

\sethlcolor{gray}

\newcommand{\bi}{\begin{itemize}}
\newcommand{\ei}{\end{itemize}}

\newtheorem*{theorem*}{Theorem}

\usepackage{amssymb}%
\usepackage{pifont}%
\newcommand{\xmark}{\ding{55}}%

\allowdisplaybreaks

\usepackage{amsmath}
\usepackage{amssymb}
\usepackage{graphicx}
\usepackage{mathtools}
\usepackage{enumerate}
\usepackage{enumitem}
\usepackage{footnote}
\usepackage{float}
\usepackage{xspace}
\usepackage{multirow}
\usepackage{nicefrac}
\usepackage{wrapfig}
\usepackage{framed}
\usepackage{url}
\usepackage{tikz}
\usetikzlibrary{arrows,chains,matrix,positioning,scopes}

\DeclareFontFamily{OMX}{MnSymbolE}{}
\DeclareFontShape{OMX}{MnSymbolE}{m}{n}{
    <-6>  MnSymbolE5
   <6-7>  MnSymbolE6
   <7-8>  MnSymbolE7
   <8-9>  MnSymbolE8
   <9-10> MnSymbolE9
  <10-12> MnSymbolE10
  <12->   MnSymbolE12}{}
\DeclareSymbolFont{mnlargesymbols}{OMX}{MnSymbolE}{m}{n}
\SetSymbolFont{mnlargesymbols}{bold}{OMX}{MnSymbolE}{b}{n}
\DeclareMathDelimiter{\llangle}{\mathopen}{mnlargesymbols}{'164}{mnlargesymbols}{'164}
\DeclareMathDelimiter{\rrangle}{\mathclose}{mnlargesymbols}{'171}{mnlargesymbols}{'171}

\usepackage{prettyref}

\newcommand{\savehyperref}[2]{\texorpdfstring{\hyperref[#1]{#2}}{#2}}
\newrefformat{eq}{\savehyperref{#1}{Eq.~\eqref{#1}}}
\newrefformat{eqn}{\savehyperref{#1}{Equation~\eqref{#1}}}
\newrefformat{lem}{\savehyperref{#1}{Lemma~\ref*{#1}}}
\newrefformat{def}{\savehyperref{#1}{Definition~\ref*{#1}}}
\newrefformat{line}{\savehyperref{#1}{line~\ref*{#1}}}
\newrefformat{thm}{\savehyperref{#1}{Theorem~\ref*{#1}}}
\newrefformat{corr}{\savehyperref{#1}{Corollary~\ref*{#1}}}
\newrefformat{cor}{\savehyperref{#1}{Corollary~\ref*{#1}}}
\newrefformat{col}{\savehyperref{#1}{Corollary~\ref*{#1}}}
\newrefformat{sec}{\savehyperref{#1}{Section~\ref*{#1}}}
\newrefformat{app}{\savehyperref{#1}{Appendix~\ref*{#1}}}
\newrefformat{assum}{\savehyperref{#1}{Assumption~\ref*{#1}}}
\newrefformat{ex}{\savehyperref{#1}{Example~\ref*{#1}}}
\newrefformat{fig}{\savehyperref{#1}{Figure~\ref*{#1}}}
\newrefformat{tab}{\savehyperref{#1}{Table~\ref*{#1}}}
\newrefformat{medpe}{\savehyperref{#1}{Algorithm~\ref*{#1}}}
\newrefformat{alg}{\savehyperref{#1}{Algorithm~\ref*{#1}}}
\newrefformat{rem}{\savehyperref{#1}{Remark~\ref*{#1}}}
\newrefformat{conj}{\savehyperref{#1}{Conjecture~\ref*{#1}}}
\newrefformat{prop}{\savehyperref{#1}{Proposition~\ref*{#1}}}
\newrefformat{proto}{\savehyperref{#1}{Protocol~\ref*{#1}}}
\newrefformat{prob}{\savehyperref{#1}{Problem~\ref*{#1}}}
\newrefformat{claim}{\savehyperref{#1}{Claim~\ref*{#1}}}
\newrefformat{que}{\savehyperref{#1}{Question~\ref*{#1}}}
\newrefformat{obs}{\savehyperref{#1}{Observation~\ref*{#1}}}

\usepackage{colortbl}
\usepackage{pgfplots}
\usetikzlibrary{calc}
\pgfplotsset{compat=1.18}
\usepackage[most]{tcolorbox}
\usepackage{pifont}
\definecolor{Ink}{HTML}{273746}
\definecolor{RuleGray}{HTML}{ADBDC9}
\definecolor{stdblue}{HTML}{477FAB}
\definecolor{ciprgreen}{HTML}{2F8179}
\definecolor{ciprwash}{HTML}{DCECE3}
\definecolor{failred}{HTML}{B96350}
\colorlet{oursrow}{ciprwash!70!white}
\newcommand{\gr}[1]{\textcolor{gray}{#1}}
\newcommand{\method}{CIPR}
\newcommand{\grpo}{GRPO}
\newcommand{\toolrl}{ToolRL (GRPO)}
\newcommand{\taskonly}{\toolrl}
\newcommand{\mitrain}{multi-interface training}
\newcommand{\Mitrain}{Multi-interface training}
\newcommand{\MI}{+ multi-interface}
\newcommand{\randonly}{\MI}
\newcommand{\plusmethod}{+ \method}
\newcommand{\basemodel}{Base}
\newcommand{\origid}{original names}
\newcommand{\canonid}{canonical names}
\newcommand{\famwm}{typos and distractor tools}
\newcommand{\Famwm}{Typos \& distractor tools}

\newcommand{\Fambt}{Cost \& latency (explicit)}
\newcommand{\famnp}{implicit cost and latency perturbations}
\newcommand{\Famnp}{Cost \& latency (implicit)}
\newcommand{\hFamwm}{\shortstack{Typos \&\\ distractor tools}}

\newcommand{\rescorekl}{rescoring consistency KL}
\newcommand{\Rescorekl}{Rescoring consistency KL}
\newcommand{\rolloutkl}{rollout consistency KL}
\newcommand{\Rolloutkl}{Rollout consistency KL}
\newcommand{\ci}[2]{#1$_{\pm #2}$}
\newcommand{\stopgrad}{\operatorname{sg}}
\newcommand{\wrongmark}{\textcolor{failred}{\xmark}}
\newcommand{\grpolabel}{\textcolor{stdblue}{\textbf{\toolrl{}}}}
\newcommand{\ciprlabel}{\textcolor{ciprgreen}{\textbf{\method{}}}}
\newcommand{\bothlabel}{\textcolor{Ink}{\textbf{Both models}}}
\newcommand{\codelink}{Code repository: \url{https://github.com/ArtinTD/cross-interface-policy-regularization}.}
\newtcolorbox{casebox}[2]{enhanced, colback=Ink!3, colbacklower=white, colframe=RuleGray, boxrule=0.5pt, arc=2pt,
    left=5pt, right=5pt, top=3pt, bottom=3pt, middle=3pt, toptitle=2pt, bottomtitle=2pt,
    fonttitle=\small\bfseries, coltitle=Ink, colbacktitle=RuleGray!30, halign title=flush left,
    title={#1\\{\mdseries\ttfamily\footnotesize #2}}, halign=flush left, halign lower=flush left,
    fontupper=\small, fontlower=\small, segmentation style={solid, draw=RuleGray, line width=0.4pt}}

\newif\ifcomments
\commentstrue

\title{One Task, Many Interfaces:
Simple Cross-Interface\\ Consistency Training
Improves LLM Agent Robustness}
\author{Artin Tajdini$^{1}$\thanks{Corresponding author: \href{mailto:artin@cs.washington.edu}{\texttt{artin@cs.washington.edu}}.} \qquad Manpreet Kaur$^{2}$ \qquad Peng Qi$^{2}$ \\[4pt]
$^{1}$University of Washington \qquad $^{2}$Uniphore}
\date{}

\begin{document}
\maketitle

\begin{abstract}

Post-training binds the competence of an LLM agent to the interface it was trained on. Semantics-preserving perturbations of that interface, such as renaming an action or changing the notation, leave the underlying task unchanged yet make the agent worse at it. Scoring each sample independently means interface randomization only broadens this dependence from a single interface to the training interface distribution, so out-of-distribution perturbations still degrade the agent. We propose Cross-Interface Policy Regularization (CIPR), a self-supervised objective that replaces action names with uninformative identifiers and presents each training sample under two interfaces in which equivalent actions are identified by construction. CIPR penalizes the KL divergence between the action distributions induced under the two interfaces, so the policy behaves consistently across them. We evaluate CIPR in both supervised and reinforcement-learning post-training, where it improves robustness. On agentic function calling, held-out interface perturbations reduce the accuracy of a post-trained agent by nearly 25\%, and CIPR recovers 10.5\% while leaving accuracy on the unchanged interface intact.\footnote{\codelink}

\end{abstract}

\section{Introduction}
\label{sec:introduction}

\begin{figure}[t]
    \centering
    \includegraphics[width=0.85\linewidth]{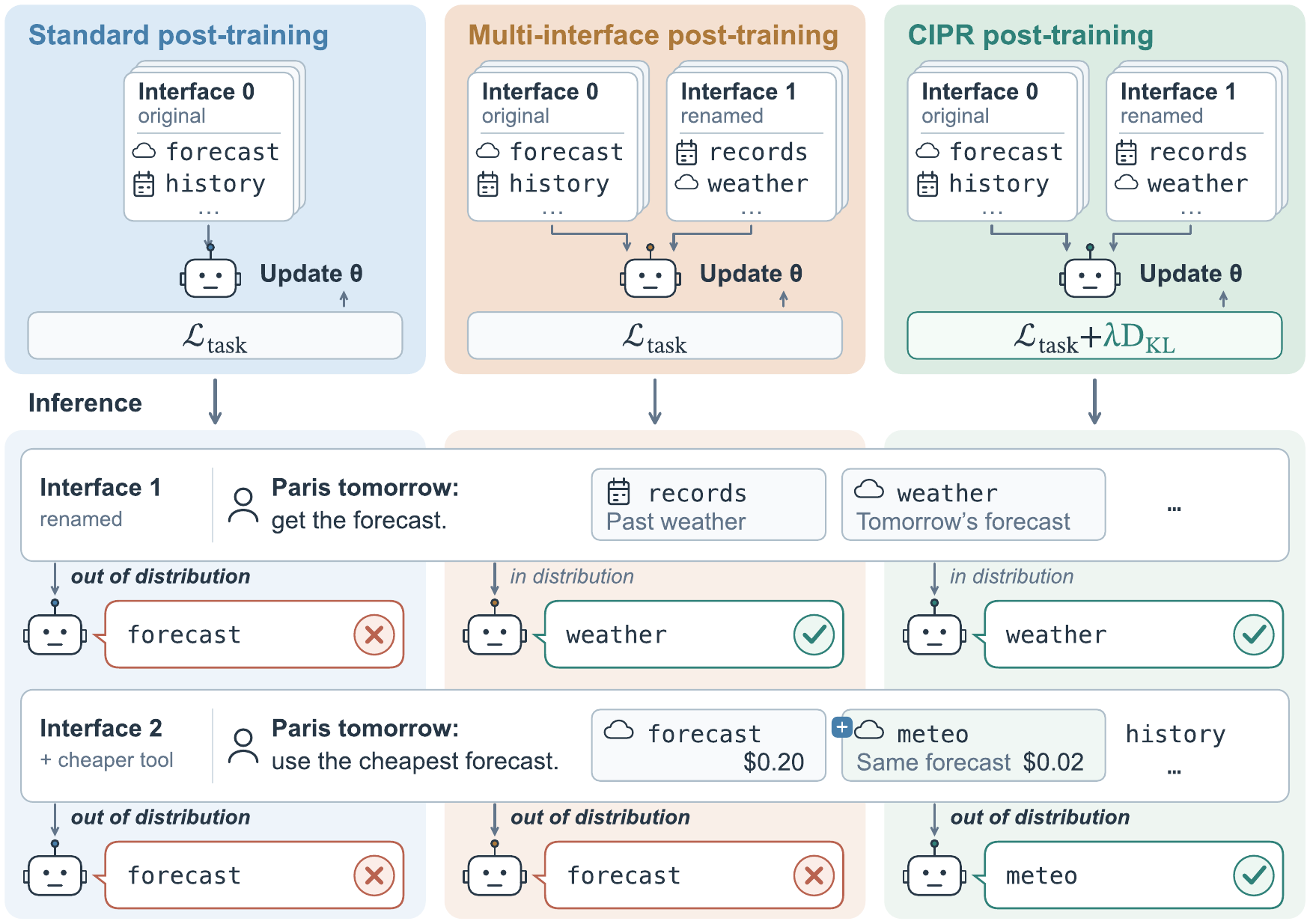}
    \caption{\textbf{Interface robustness of a tool-using agent} (schematic).
    In each column, we post-train an agent to fetch weather forecasts (top) and then test it at inference (bottom) on a renamed interface (\emph{Interface~1}) and on one with an added, cheaper forecast tool (\emph{Interface~2}).
    The label on each arrow into an agent says whether the test interface is in that agent's training distribution.
    Standard post-training on one interface fails both.
    \Mitrain{} handles the renaming, which it saw during training, but misses the new tool.
    \method{}, which also compares the policy across interfaces, chooses by description in both cases.}
    \label{fig:motivation}
\end{figure}

Language models are now trained to act as tool-using agents, either natively~\citep{qwen35,kimik2}, or by fine-tuning on agent trajectories and reinforcement learning from task rewards~\citep{zengAgentTuning2023,qian2025toolrl}.
What these agents learn, however, is tied to the interface through which their tools are presented, meaning the names, descriptions, and argument schemas of the available tools.
Figure~\ref{fig:motivation} illustrates this with a weather agent trained on a single interface, which fails when its forecast tool is renamed and when a cheaper equivalent is added, although neither change alters the task.
Agents fine-tuned on trajectories can lose most of their task success when action names are minimally rewritten~\citep{gu2026pipe}, and function-calling models degrade when tools are renamed, similar tools are added, or requests are rephrased~\citep{ye2024rotbench,rabinovich2025robustness,zhouRobustBench2026}.
Such changes are routine in deployment, where registries combine tools from different servers, providers revise their schemas, and users misspell requests~\citep{zhouRobustBench2026}.
Interfaces can also be manipulated, since edits to a tool's description alone can make agents prefer it over equivalent tools~\citep{faghih2025toolpreferences}.

This failure is an instance of shortcut learning~\citep{geirhos2020shortcut}.
On a fixed training interface, a policy that reads the tool descriptions and a policy that relies on a familiar name or a position in the tool list make the same calls and earn the same reward, so the task objective has no reason to prefer the first.
The second policy fails as soon as the interface changes, even if the task is unchanged.
Such a policy can still succeed on new tasks under its training interface, so the failure is one of generalizing across interfaces rather than of fitting noise in the training examples.

Prior training methods reduce this sensitivity by changing what the agent trains on.
Function masking replaces tool and argument names with random strings~\citep{lin2025hammer}, which removes one cue but leaves others, such as a tool's position or the set of tools offered.
\Mitrain{}, also known as domain randomization, varies tool schemas, metadata, and question formats during training~\citep{zhouRobustBench2026,liu2026formatmix}, and synthetic corpora such as ToolACE broaden the tools seen in training~\citep{liuToolACE2024}.
Both approaches still score each training interface on its own.
Their objective asks the policy to succeed under every interface it sees, but never to treat equivalent interfaces alike, so a policy can fit every training variant while keeping a shortcut that fails on a new kind of change, as in the middle column of Figure~\ref{fig:motivation}.

The interface changes an agent meets in deployment cannot all be anticipated, so robustness should not depend on training on each of them.
Instead, the agent should treat interfaces that offer equivalent tools alike, including interfaces it has never seen.
We propose \emph{Cross-Interface Policy Regularization} (\method{}), which trains for this property directly (Figure~\ref{fig:method}).
\method{} uses a correspondence that \mitrain{} discards, since whenever we construct a perturbed interface, we know which of its tools is equivalent to which tool in the original.
We canonicalize each training interface, replacing every tool name with a neutral identifier such as \texttt{function\_01}, so that names carry no task information.
Canonicalization leaves other cues in place, such as a tool's position, so we pair the canonical interface with a perturbed copy in which the tools are renamed, reordered, and joined by degraded twins.
We then penalize the KL divergence between the agent's action distributions under the two interfaces, with the canonical interface as the reference.
Unlike prior consistency regularization, which asks a model for the same output under both inputs~\citep{xie2020uda,zhou2026pirl}, this term lets the response change with the tool names and compares only the distributions over actions, with each tool matched to its counterpart through the known correspondence.
The consistency KL has two terms.
The \emph{\rescorekl{}} holds a response fixed and compares the action distributions along it, which needs no additional generation, and the \emph{\rolloutkl{}} lets each interface produce its own responses, since a different interface can call for different reasoning.
We add both terms to the task loss, and they need no labels beyond those of the task.

We evaluate \method{} in reinforcement-learning and supervised post-training under interface perturbations held out from training.
On function calling with two backbones, \method{} is more robust than the base model and standard post-training, and it improves on \mitrain{} in all three groups of held-out perturbations while keeping original accuracy (Table~\ref{tab:main}).

Our contributions are as follows.
\begin{itemize}
    \item \method{}, a self-supervised objective that penalizes the KL divergence between an agent's action distributions under two interfaces whose equivalent actions are known by construction, with a \rescorekl{} that needs no additional generation and a \rolloutkl{} that lets the agent's reasoning differ between the interfaces (Section~\ref{sec:method}).
    \item A tool-calling instantiation that canonicalizes tool names and pairs every training interface with a perturbed copy in which only descriptions and schemas identify the right tool (Sections~\ref{sec:canon} and~\ref{sec:perturb}).
    \item Experiments in reinforcement-learning and supervised post-training, in which \method{} improves on standard post-training and \mitrain{} under held-out interface perturbations and makes an agent fine-tuned on trajectories robust to renamed actions (Section~\ref{sec:experiments}).
\end{itemize}

\section{Related Work}
\label{sec:related}

\paragraph{Interface sensitivity of tool-using agents.}
Tool-use benchmarks mostly evaluate agents on the interface released with them~\citep{patil2025bfcl,barres2025tau2}, yet which tool an agent calls depends on how the tools are presented.
Renaming or corrupting tool and argument names, rephrasing queries, and adding related tools all lower accuracy~\citep{ye2024rotbench,rabinovich2025robustness,zhouRobustBench2026}, the wording and order of equivalent tools change which one is selected~\citep{faghih2025toolpreferences,blankenstein2026biasbusters}, and agents call a tool whose parameters fit the query even when the tool does not serve it~\citep{liu2026sabeval}.
Similar sensitivity to semantics-preserving rewrites appears in mathematical reasoning~\citep{mirzadeh2025gsmsymbolic} and in choices among reordered options~\citep{egressy2025setllm}.
Training can deepen this reliance: PIPE, which rewrites environment interfaces while preserving task semantics, finds that fine-tuning on trajectories increases reliance on the training interface~\citep{gu2026pipe}, and agent-tuning data entangles format following with reasoning~\citep{chen2024agentflan}.
These studies measure the problem, whereas we address it during training.

\paragraph{Training on varied interfaces.}
Most remedies applied during training change what the agent sees.
They hide a cue, as function masking replaces names with random strings~\citep{lin2025hammer,qian2025toolrl}; broaden the tools and environments seen in training, through synthesized APIs and agent data~\citep{liuToolACE2024,zhang2025xlam,he2025gentool,mekala2024toolverifier}, generated environments~\citep{sullivan2025randomworld,du2026codegym}, and tools that change over time~\citep{chen2025toolevo}; or pair clean and perturbed versions of the same tasks~\citep{lv2026openagent,chen2026noisyagent}.
Closest to our setting, ToolRL-DR randomizes the interface during tool-use reinforcement learning~\citep{zhouRobustBench2026}, as multi-format supervision does for question formats~\citep{liu2026formatmix}.
Others change the target instead, scoring an action by its effect rather than its exact string~\citep{tang2026asl}.
Outside agents, shortcuts of this kind persist after the core features are learned~\citep{qiu2024complexity}, and regularization or data rebalancing aimed at them can suppress causal features or lower worst-group accuracy~\citep{hong2025regularization,labonte2024grouprobustness}.
In all of these, the objective still scores each training example on its own.
We keep the perturbed interfaces and add a term that compares the policy across them.

\paragraph{Consistency across equivalent inputs.}
Consistency regularization penalizes the divergence between a model's predictions on two versions of an input that should receive the same prediction, such as an example and its augmentation in semi-supervised learning~\citep{xie2020uda} or semantically equivalent pairs in domain generalization~\citep{gao2024lam}.
It has been applied to reinforcement-learning policies on augmented observations~\citep{raileanu2021drac} and to vision-language-action policies under perturbed instructions and observations~\citep{luo2026rovla}.
In concurrent work, PIRL applies it to reinforcement learning of multimodal LLMs with a per-token KL between the policy's next-token distributions for the same response under a clean and an adversarially perturbed prompt~\citep{zhou2026pirl}.
Each of these methods compares the two predictions directly, which assumes the correct output is the same under both inputs.
Renaming tools breaks this assumption, because the correct call is then written with different tokens, so we compare distributions over actions matched through the known correspondence between the two interfaces.

\section{Cross-Interface Policy Regularization}
\label{sec:method}

\begin{figure}[t]
    \centering
    \includegraphics[width=0.85\linewidth]{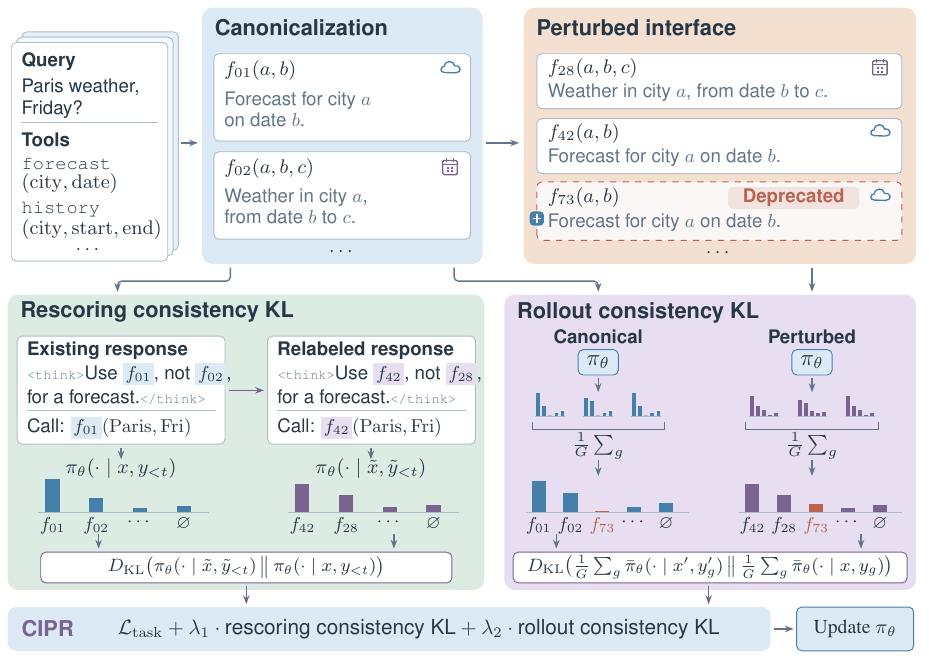}
    \caption{\textbf{Overview of \method{}.}
    We first canonicalize the interface, renaming the two tools of a weather query $f_{01}$ and $f_{02}$, short for \texttt{function\_01} and \texttt{function\_02}.
    We then build a perturbed interface that adds a degraded twin of every tool, such as the deprecated forecast $f_{73}$, and assigns all tools new identifiers in a new order.
    Each bar chart is an action distribution, the probability of calling each tool or none ($\varnothing$).
    The \rescorekl{} holds one response fixed and compares these distributions at each of its decisions, before and after its tool names are relabeled in the reasoning and the call.
    The \rolloutkl{} lets each interface produce its own responses, so the reasoning can differ, and compares their average distributions, in which the twin $f_{73}$ has probability zero under the canonical interface.
    Probabilities are illustrative.}
    \label{fig:method}
\end{figure}

We consider an LLM agent that receives a user query together with an interface, which lists the available tools with their names, descriptions, and argument schemas.
At each step the agent either calls one of the listed tools or responds without a call.
A perturbation of the interface is semantics-preserving when the tool that solves the task is still available and still solves it, even though its name, its position, or the surrounding options have changed.
We want the agent to assign the same probability to equivalent tools under such perturbations, and \method{} trains for this in three steps.
We canonicalize the interface so that tool names carry no task information (Section~\ref{sec:canon}), pair every training example with a perturbed copy of its interface (Section~\ref{sec:perturb}), and penalize the divergence between the agent's action distributions under the two interfaces (Section~\ref{sec:ckl}).
Figure~\ref{fig:method} summarizes the procedure.

\subsection{Canonicalization}
\label{sec:canon}

The most direct shortcut available to a trained agent is a familiar tool name.
Because a name that led to correct calls during training usually still does so at test time, a policy can select tools by name without reading their descriptions, and function-calling models are known to be misled by naming conventions~\citep{lin2025hammer}.
We remove this shortcut by canonicalizing the interface, replacing every tool name with a neutral identifier and every argument name with an identifier numbered within its tool, so that \texttt{forecast(city, date)} becomes \texttt{function\_01(arg\_01, arg\_02)}.
We keep descriptions, types, required arguments, and argument values unchanged, so the agent has to read the descriptions to tell the tools apart, and we apply the same mapping to every reference to these names in the query, in earlier turns, and in the training targets.
At inference, a canonicalization adapter applies this mapping to the incoming interface and translates the generated call back to the executable API, so the deployed tools themselves are never modified.
Canonicalization is related to function masking~\citep{lin2025hammer}, a training strategy that replaces names with random strings so that the model attends to descriptions; unlike masking, we also apply the mapping at inference, so the policy never meets the original names.

\subsection{Perturbed interfaces}
\label{sec:perturb}

The consistency KL of Section~\ref{sec:ckl} needs only a perturbed interface in which we know which tool is equivalent to which canonical tool.
We design ours to also remove the cues that canonicalization leaves in place, such as a tool's position in the tool list or the identifiers seen during training, and construct it for every training example from the canonical interface in three steps.
We first give every tool a degraded twin (\textbf{twin expansion}), a copy whose description states that it is worse, for example deprecated or slower, or whose schema replaces an argument the task needs with an unrelated one.
We then give the expanded tool list new identifiers (\textbf{relabeling}) and shuffle the tools and their arguments (\textbf{reordering}).
The query and the descriptions of the original tools are unchanged, and each twin differs from its original only in its description or schema.
Appendix~\ref{app:twins} lists the exact operations.

\subsection{Consistency KL on equivalent actions}
\label{sec:ckl}

The task objective never compares which tools the policy chooses under the canonical and the perturbed interface.
We add this comparison by building on consistency regularization~\citep{xie2020uda,raileanu2021drac}, which penalizes the divergence between a model's predictions on two versions of an input that call for the same output.
An agent's correct response, however, changes with the interface, because the tools it calls have different names, and what stays fixed is the correspondence between equivalent tools, which is known from the construction of the perturbed interface.
We therefore compare the policy's distributions over actions rather than over tokens.

\paragraph{Action distributions.}
The policy $\pi_\theta$ generates a response $y$ to a prompt $x$, which contains the interface, the query, and any earlier turns.
Within $y$, the policy makes a decision each time it begins a tool call or begins the text that ends its turn, and each decision is an action $a$, either a tool declared in $x$ or $\varnothing$, answering without a call.
Writing $y_{<t}$ for the part of $y$ before its $t$-th decision, we denote by $\pi_\theta(a \mid x, y_{<t})$ the probability that the policy continues $y_{<t}$ with action $a$.
We read this probability in one forward pass from the probability of the tokens that open a call and of each declared tool's name after them, normalized over the declared tools (Appendix~\ref{app:training}), as in work that reads the action probabilities of LLM agents from their token probabilities~\citep{carta2023glam,tan2024twosome}.
The two interfaces name the same tool differently, so we index actions by equivalence rather than by name or position, and a tool and its counterpart under the other interface share an index.
A tool that only one interface declares, such as a twin, has an index of its own, with probability zero under the interface that does not declare it, so probability moved onto such a tool counts as disagreement instead of being merged with another.
Arguments do not enter this distribution and are trained by the task objective alone.

\paragraph{Two consistency terms.}
We compare these distributions under the two interfaces in two ways, which differ in whether the agent's reasoning is held fixed (Figure~\ref{fig:method}).
\begin{itemize}
    \item \textbf{\Rescorekl{}.}
    When only the tool names change, the reasoning behind a response remains valid under both interfaces, so this term holds a response fixed and compares the decisions along it.
    We take a response $y$ to the canonical prompt $x$, a rollout of the task objective or a demonstration in supervised post-training, and relabel it, giving every tool a new neutral name in the prompt and throughout the response.
    Each decision of the relabeled response $\tilde y$ to the relabeled prompt $\tilde x$ sits at the same position as in $y$, and we average the divergence over the $n_y$ decisions of $y$,
    \begin{equation}
        \mathrm{KL}_{\mathrm{rescore}}(x, y) = \frac{1}{n_y} \sum_{t=1}^{n_y} D_{\mathrm{KL}}\big( \pi_\theta(\cdot \mid \tilde x, \tilde y_{<t}) \,\big\|\, \stopgrad[\pi_\theta(\cdot \mid x, y_{<t})] \big).
        \label{eq:rescore}
    \end{equation}
    Scoring $\tilde y$ needs only forward passes and no generation.
    \item \textbf{\Rolloutkl{}.}
    When the interface changes which tools are available, the agent should reason differently, and comparing its decisions along a fixed response would penalize exactly that.
    This term therefore lets each interface produce its own responses.
    We sample $G$ responses $y_1, \dots, y_G$ to $x$ and $G$ responses $y'_1, \dots, y'_G$ to $x'$, the prompt with the full perturbed interface of Section~\ref{sec:perturb}, and compare the average action distributions of the two groups,
    \begin{equation}
        \mathrm{KL}_{\mathrm{rollout}}(x) = D_{\mathrm{KL}}\Big( \frac{1}{G} \sum_{g=1}^{G} \bar\pi_\theta(\cdot \mid x', y'_g) \,\Big\|\, \stopgrad\Big[ \frac{1}{G} \sum_{g=1}^{G} \bar\pi_\theta(\cdot \mid x, y_g) \Big] \Big),
        \label{eq:rollout}
    \end{equation}
    where $\bar\pi_\theta(a \mid x, y) = \frac{1}{n_y} \sum_{t=1}^{n_y} \pi_\theta(a \mid x, y_{<t})$ averages the action distribution over the decisions of one response.
    Only the responses are sampled, so each rollout contributes its whole action distribution at every decision rather than only the action it took.
    Gradients flow through these probabilities but not through the sampling.
    Because the two groups are sampled independently, their responses do not correspond, so we compare only their averages, at the cost of $G$ additional rollouts per prompt.
\end{itemize}

\paragraph{Objective.}
We add both terms to the task loss of Section~\ref{sec:objective},
\begin{equation}
    \mathcal{L}(\theta) = \mathcal{L}_{\mathrm{task}}(\theta) + \lambda_1\, \mathbb{E}_{x, y}\big[ \mathrm{KL}_{\mathrm{rescore}}(x, y) \big] + \lambda_2\, \mathbb{E}_{x}\big[ \mathrm{KL}_{\mathrm{rollout}}(x) \big],
    \label{eq:objective}
\end{equation}
where $x$ ranges over training prompts and $y$ over the responses that the task loss uses, $D_{\mathrm{KL}}(p \,\|\, q) = \sum_{a} p(a) \log \frac{p(a)}{q(a)}$ sums over the tools and $\varnothing$, $\stopgrad$ stops the gradient, and $\lambda_1, \lambda_2 \geq 0$.
In both terms the reference is the policy itself under the canonical interface, so no teacher model or additional labels are needed.
We stop its gradient so that the perturbed distribution moves toward it rather than both moving toward each other.
A tool that only the perturbed interface declares has reference probability zero, so we floor reference probabilities at a small constant $\epsilon$ inside the logarithm, which keeps the divergence finite and makes the penalty for such a tool grow with its probability under the perturbed interface.

\subsection{Task loss and inference}
\label{sec:objective}

The consistency terms do not depend on how $\mathcal{L}_{\mathrm{task}}$ is computed, and we use two task losses.
In reinforcement learning, we use the \grpo{} objective of ToolRL~\citep{qian2025toolrl}, computed on rollouts under both interfaces.
In supervised post-training, we use a distillation loss and only the \rescorekl{} (Section~\ref{sec:sft}).
We add the consistency terms to the loss rather than to the reward, so agreement cannot substitute for correct calls and the task loss remains responsible for solving the task.
At inference, we run the model with the canonicalization adapter alone; perturbed interfaces, relabeled copies, and paired rollouts exist only during training.
Algorithm~\ref{alg:cipr} in Appendix~\ref{app:training} gives the full procedure.

Nothing in the consistency KL is specific to tools; it requires only actions with model probabilities and a perturbation under which equivalent actions are known.

\section{Experiments}
\label{sec:experiments}

We evaluate \method{} in reinforcement-learning and supervised post-training of agents.
If the consistency KL removes interface shortcuts rather than fitting the training perturbations, \method{} should raise accuracy under held-out perturbations, keep or raise accuracy on the original interface, and improve on \mitrain{}, which sees the same perturbed interfaces without the consistency KL.
We test these three predictions by adding \method{} to reinforcement learning for function calling (Section~\ref{sec:main}); we then apply it to supervised post-training on WebShop (Section~\ref{sec:sft}) and ablate its components (Section~\ref{sec:ablations}).

\subsection{Setup}
\label{sec:setup}

We build all reinforcement-learning experiments on the public ToolRL recipe~\citep{qian2025toolrl}, so the compared models differ only in their training interfaces and objective.

\paragraph{Training.}
We train with \grpo{}~\citep{shao2024deepseekmath} on the $3{,}920$ ToolRL training examples, rewarding correct calls and format.
We use two backbones, Qwen2.5-3B-Instruct~\citep{qwen25} with tools in the prompt and Qwen3.5-4B~\citep{qwen35} with thinking and native function calling.
Appendix~\ref{app:training} lists hyperparameters.

\paragraph{Baselines.}
We compare four models on each backbone.
\emph{\basemodel{}} is the backbone without post-training, and \emph{\toolrl{}}, our standard post-training baseline, is the backbone post-trained with the ToolRL recipe; both are evaluated with the original tool names, as they would be deployed.
\Mitrain{} adds the perturbed interfaces of Section~\ref{sec:perturb} to ToolRL training, as in ToolRL-DR~\citep{zhouRobustBench2026}, and \method{} adds both consistency terms to it; the tables label these models \MI{} and \plusmethod{}.
We train both with canonical names on the same interfaces and evaluate both through the canonicalization adapter, so that they differ only in whether the policy is compared across them.

\paragraph{Held-out perturbations.}
We evaluate on the $2{,}501$ single-turn items of BFCL~v4~\citep{patil2025bfcl} that have a reference call, under thirteen perturbation types adapted from RobustBench-TC~\citep{zhouRobustBench2026} (Appendix~\ref{app:eval}).
None of them is used during training, and we report them in three groups.
\emph{\Famwm{}} introduces typos into the query or adds a distractor for every tool the task calls, either an unrelated tool renamed to resemble it or a duplicate with the same name whose description or parameters are corrupted.
The six cost and latency types add a cheaper or faster equivalent of the correct tool under a new name, so that the familiar name now points to the worse choice.
As in RobustBench-TC, the query in \emph{\Fambt{}} asks for the better tool, for example ``Please use a cost-effective option''. This request tells the agent what to prefer, so we add a harder version, \emph{\Famnp{}}, in which the query only asks the agent to describe the available tools and states no preference.
These two groups are the hardest test, because the right choice depends only on a stated cost or latency, and preferring a cheaper or faster tool is never rewarded during training.
Beyond single-turn calls, we evaluate the three groups on the airline and retail domains of $\tau^2$-bench~\citep{barres2025tau2}, where the agent completes multi-turn tasks with a simulated user, and Appendix~\ref{app:rotbench} also reports RoTBench~\citep{ye2024rotbench}, which corrupts tool and parameter names at graded levels.

\subsection{Robustness to held-out perturbations}
\label{sec:main}

Table~\ref{tab:main} tests the three predictions on BFCL.

\begin{table}[t]
    \centering
    \caption{\textbf{Full-call accuracy (\%) on BFCL under interface perturbations.}
    \emph{Tool names} gives the names each model sees at evaluation.
    Each model is evaluated with the names it was trained on, as it would be deployed, so \basemodel{} and \toolrl{} see the original names, and the multi-interface and \method{} models, which are trained with canonical names, see canonical names through the canonicalization adapter.
    Gray rows evaluate \basemodel{} and \toolrl{} through the canonicalization adapter as well.
    \emph{In distribution} applies the perturbation used in training (Section~\ref{sec:perturb}), which is defined on canonical names, so rows with the original names show n/a.
    \emph{Mean} averages the three held-out groups, and \emph{Gap} is original accuracy minus the mean.
    Bold marks the best value per backbone, excluding gray rows.
    All $95\%$ confidence intervals have half-widths of at most $2.0$ points, and Appendix~\ref{app:extra} reports each perturbation type.}
    \label{tab:main}
    \small
    \setlength{\tabcolsep}{2.6pt}
    \begin{tabular}{llccccccc}
        \toprule
        & & & \multicolumn{1}{c}{In distribution $\uparrow$} & \multicolumn{4}{c}{Held-out perturbations $\uparrow$} & \\
        \cmidrule(lr){4-4}\cmidrule(lr){5-8}
        Model & \shortstack{Tool\\ names} & Original $\uparrow$ & \shortstack{Twins \&\\ reordering} & \shortstack{Typos \&\\ distractors} & \shortstack{Cost \& lat.\\ (explicit)} & \shortstack{Cost \& lat.\\ (implicit)} & Mean & Gap $\downarrow$ \\
        \midrule
        \multicolumn{9}{l}{\textit{Qwen2.5-3B, tools in prompt}} \\
        \basemodel{} & original & 63.7 & n/a & 59.9 & 24.2 & 17.7 & 33.9 & 29.8 \\
        & \gr{canonical} & \gr{61.5} & \gr{43.5} & \gr{59.3} & \gr{41.9} & \gr{28.5} & \gr{43.2} & \gr{18.3} \\
        \toolrl{} & original & 73.3 & n/a & 71.6 & 29.3 & 20.9 & 40.6 & 32.7 \\
        & \gr{canonical} & \gr{70.7} & \gr{49.1} & \gr{68.9} & \gr{46.0} & \gr{33.5} & \gr{49.5} & \gr{21.2} \\
        \MI{} & canonical & 74.0 & 68.2 & 71.5 & 51.5 & 39.6 & 54.2 & 19.8 \\
        \rowcolor{oursrow} \textbf{\plusmethod{} (ours)} & canonical & \textbf{74.9} & \textbf{68.8} & \textbf{72.2} & \textbf{53.6} & \textbf{42.3} & \textbf{56.1} & \textbf{18.8} \\
        \midrule
        \multicolumn{9}{l}{\textit{Qwen3.5-4B, native function calling}} \\
        \basemodel{} & original & 84.2 & n/a & 83.4 & 83.1 & 55.3 & 73.9 & 10.2 \\
        & \gr{canonical} & \gr{82.0} & \gr{76.3} & \gr{81.3} & \gr{81.0} & \gr{58.8} & \gr{73.7} & \gr{8.3} \\
        \toolrl{} & original & 84.8 & n/a & 84.2 & \textbf{84.8} & 60.2 & 76.4 & 8.4 \\
        & \gr{canonical} & \gr{84.6} & \gr{77.1} & \gr{84.0} & \gr{84.0} & \gr{52.3} & \gr{73.4} & \gr{11.1} \\
        \MI{} & canonical & 84.8 & \textbf{84.8} & 84.1 & 84.2 & 70.7 & 79.7 & 5.1 \\
        \rowcolor{oursrow} \textbf{\plusmethod{} (ours)} & canonical & \textbf{85.0} & 83.8 & \textbf{84.5} & 84.6 & \textbf{73.0} & \textbf{80.7} & \textbf{4.2} \\
        \bottomrule
    \end{tabular}
\end{table}

\paragraph{Comparison with \toolrl{}.}
Standard post-training leaves most of the gap in place.
On Qwen2.5-3B, \toolrl{} raises original accuracy by $9.6$ points, yet its gap grows from $29.8$ to $32.7$ points, because the familiar name it learned to follow now points to the worse tool.
On Qwen3.5-4B, whose base model is already trained for tool use, the gap narrows only from $10.2$ to $8.4$ points.

\paragraph{Canonical names alone.}
Canonical names are not a free improvement for a model trained without them (gray rows).
They lower the original accuracy of both base models and of \toolrl{}, and although they raise the cost and latency accuracy of \toolrl{} on Qwen2.5-3B, they lower its accuracy in every held-out group on Qwen3.5-4B.

\paragraph{Comparison with \mitrain{}.}
Training on perturbed interfaces makes canonical names pay off, and the consistency KL adds to that gain.
On both backbones, \method{} is higher than \mitrain{} on the original interface and in every held-out group.
The largest gains are on the cost and latency perturbations, the ones least like anything seen in training, with $2.1$ and $2.7$ points on Qwen2.5-3B and $2.3$ points on the implicit version for Qwen3.5-4B, while the other differences are within the $95\%$ confidence intervals.
Under the perturbation used in training (\emph{In distribution}), the two models differ by about a point or less on both backbones, so the advantage of \method{} lies in the perturbations that training did not cover.
On both backbones, \method{} thus matches or exceeds the original accuracy of \toolrl{} while raising its mean accuracy under perturbation by $15.5$ and $4.3$ points.

\paragraph{Interactive tasks.}
On $\tau^2$-bench, the Qwen3.5-4B \method{} model has the smallest gap in both domains between its original reward and its mean reward under the three groups, although \toolrl{} with the original names has a higher reward on the original interface and under every group (Table~\ref{tab:tau2}).

\subsection{Supervised post-training}
\label{sec:sft}

We next apply \method{} to an agent trained by supervised learning, AgentLM-7B~\citep{zengAgentTuning2023}, whose WebShop policy collapses when its actions are renamed~\citep{gu2026pipe}.

\paragraph{Setting.}
AgentLM is fine-tuned on agent trajectories mixed with general instruction data, and acts on WebShop~\citep{yaoWebShop2022} through two actions named \texttt{search} and \texttt{click}.
The released model already solves the task under its original action names, so we apply \method{} as self-distillation.
The released model is the teacher and initializes the student, which we train to match it under canonical names and, through the \rescorekl{}, under relabeled names, without additional rollouts.

\begin{figure}[t]
    \begin{minipage}[b]{0.49\linewidth}
    \centering
    \begin{tikzpicture}
    \begin{axis}[
        xbar, scale only axis, width=3.1cm, height=4.7cm, bar width=2.9pt,
        xmin=0, xmax=100, xtick={0,50,100}, axis x line*=bottom, axis y line*=left,
        xlabel={Mean reward ($\times 100$)}, xlabel style={font=\footnotesize},
        symbolic y coords={Swapped,Random,Opaque,Suffix,Synonym,Original,Canonical},
        ytick=data, major tick length=0pt, tick label style={font=\footnotesize},
        y tick label style={align=right},
        yticklabels={{Swapped\\[-2pt]{\scriptsize\texttt{click}, \texttt{search}}}, {Random\\[-2pt]{\scriptsize $100$ characters}}, {Opaque\\[-2pt]{\scriptsize\texttt{z1}, \texttt{z2}}}, {Synonyms with suffix\\[-2pt]{\scriptsize\texttt{find\_\_q1}, \texttt{press\_\_q2}}}, {Synonyms\\[-2pt]{\scriptsize\texttt{find}, \texttt{press}}}, {Original\\[-2pt]{\scriptsize\texttt{search}, \texttt{click}}}, {\textbf{Canonical}\\[-2pt]{\scriptsize\texttt{function\_1}, \texttt{function\_2}}}},
        enlarge y limits=0.07, clip=false,
        nodes near coords, nodes near coords style={font=\tiny, /pgf/number format/.cd, fixed, precision=1, zerofill},
    ]
    \addplot[fill=ciprgreen, draw=none] coordinates {(0.0,Swapped) (66.9,Random) (81.4,Opaque) (79.6,Suffix) (76.0,Synonym) (81.6,Original) (81.0,Canonical)};
    \addplot[fill=stdblue, draw=none] coordinates {(0.0,Swapped) (63.9,Random) (78.1,Opaque) (7.5,Suffix) (5.5,Synonym) (78.9,Original) (75.7,Canonical)};
    \draw[RuleGray, densely dashed] ($(axis cs:0,Canonical)!0.5!(axis cs:0,Original)$) -- ($(axis cs:100,Canonical)!0.5!(axis cs:100,Original)$);
    \end{axis}
    \end{tikzpicture}
    \captionof{figure}{\textbf{WebShop} evaluation on AgentLM-7B before (blue) and after (green) \method{} self-distillation, under different names.
    }
    \label{fig:sft}
    \end{minipage}\hfill
    \begin{minipage}[b]{0.49\linewidth}
    \centering
    \captionof{table}{\textbf{Ablations on Qwen2.5-3B} (accuracy, \%, canonical names).
    Each row removes one component from \method{}, and removing the consistency KL gives \mitrain{}; $^\dagger$one added near-duplicate tool takes the place of the twins.
    The gray row is \toolrl{} with canonical names.}
    \label{tab:ablation}
    \footnotesize
    \setlength{\tabcolsep}{2pt}
    \begin{tabular}{@{}lcc@{}}
        \toprule
        & & Cost \& \\
        & & latency \\
        & Original $\uparrow$ & (explicit) $\uparrow$ \\
        \midrule
        \rowcolor{oursrow} \plusmethod{} & 74.9 & 53.6 \\
        $-$ consistency KL & 73.5 & 51.7 \\
        $-$ twins & 73.5 & 48.1 \\
        $-$ twins and reordering$^\dagger$ & 74.7 & 43.7 \\
        \midrule
        \gr{\toolrl{}} & \gr{70.7} & \gr{46.0} \\
        \bottomrule
    \end{tabular}
    \vspace{12pt}
    \end{minipage}
\end{figure}

\paragraph{Results.}
Consistent with \citet{gu2026pipe}, renaming \texttt{search} and \texttt{click} to the synonyms \texttt{find} and \texttt{press} reduces the released model's reward from $78.9$ to $5.5$ (Figure~\ref{fig:sft}).
After self-distillation, the agent reaches $81.0$ under canonical names, the interface it is deployed with, against $75.7$ for the released model behind the same adapter, so the adapter alone does not explain the gain.
Seeing the names directly, the trained agent is also robust to neutral renaming, with $76.0$ under the synonyms.
Swapped names, each of which suggests the other action, drive both models to zero, which is why \method{} keeps the canonicalization adapter, behind which a misleading name never reaches the policy.

\subsection{Ablations}
\label{sec:ablations}

Table~\ref{tab:ablation} ablates the consistency KL, the twins, and the reordering of the perturbed interface on Qwen2.5-3B, with every model evaluated under canonical names (Table~\ref{tab:ablation-full} lists every run).
It reports accuracy on the original interface and on the explicit cost and latency perturbations, where only the stated cost or latency identifies the right tool.

\paragraph{Effect of the consistency KL.}
Without the consistency KL, the method reduces to \mitrain{} on the same interfaces, and accuracy falls from $53.6\%$ to $51.7\%$ on the explicit cost and latency perturbations and from $74.9\%$ to $73.5\%$ on the original interface, consistent with the difference between the two methods in Table~\ref{tab:main}.

\paragraph{Effect of the perturbed interface.}
The twins contribute the most.
Without them, accuracy on the explicit cost and latency perturbations falls from $53.6\%$ to $48.1\%$, and with a single added near-duplicate tool in place of twins and reordering it falls to $43.7\%$, below \toolrl{} with canonical names.
Trained with a single added tool, the model relies on position rather than on the stated costs, with $70.8\%$ when the correct tool comes first and $16.5\%$ when the costlier equivalent does (Table~\ref{tab:ablation-full}).
The consistency KL also gains more from twins, adding $1.9$ points with them against $0.6$ with reordering alone (Table~\ref{tab:ablation-full}).

\paragraph{Effect of canonicalization at inference.}
Canonical names are needed at inference as well.
Evaluated with the original names, the \method{} model reaches $28.7\%$ on the explicit cost and latency perturbations in the original tool order, against $60.5\%$ with canonical names, because the familiar name belongs to the costlier tool (Appendix~\ref{app:extra}).

\section{Conclusion}
\label{sec:conclusion}

LLM agents learn to use tools through a particular interface, and the task objective cannot tell a decision that depends on the task from one that depends on how the tools are presented.
We supply that distinction with \method{}, a self-supervised consistency KL between the agent's action distributions under two interfaces whose equivalent actions are known by construction.
Added to \mitrain{}, it improves accuracy in all three groups of held-out BFCL perturbations on both backbones without lowering accuracy on the original interface (Table~\ref{tab:main}).
Applied as self-distillation, it makes AgentLM robust to renaming its WebShop actions.
Because the targets come from the policy itself, the objective needs no annotation beyond the task, which makes it a natural candidate for agentic training at scale.

\paragraph{Limitations.}
\method{} needs the canonicalization adapter at inference, because a name that misrepresents its action, such as the swapped WebShop names in Figure~\ref{fig:sft}, still misleads a trained agent that sees it directly.
The consistency KL constrains which tool the agent calls and whether it calls one, while arguments are learned from the task loss alone, and argument errors barely change in our error analysis (Appendix~\ref{app:errors}).
It also requires a perturbation whose equivalences are known, which is direct for tool lists but harder to construct for richer actions such as edits to a program.
The \rolloutkl{} samples an additional group of rollouts per prompt, which raises the cost of training.
Our experiments use models of up to $9$B parameters, one training run per configuration, and mostly single-turn function calling, so the effect at larger scale and in long multi-turn tasks remains to be established.

\bibliographystyle{plainnat}
\bibliography{references}

\clearpage
\appendix

\section{Training details}
\label{app:training}

This appendix gives the hyperparameters of the reinforcement-learning runs (Table~\ref{tab:hparams}) and the full \method{} training procedure (Algorithm~\ref{alg:cipr}).

\paragraph{Hyperparameters.}
All reinforcement-learning runs build on ToolRL~\citep{qian2025toolrl} and keep its optimizer settings.
The task objective is \grpo{}~\citep{shao2024deepseekmath} with rewards for call correctness and response format and no reference-policy KL; calling a degraded twin earns the same reward as any other wrong call.

\begin{table}[!ht]
    \centering
    \caption{Hyperparameters of the reinforcement-learning runs.}
    \label{tab:hparams}
    \small
    \begin{tabular}{@{}lcc@{}}
        \toprule
        & Qwen2.5-3B & Qwen3.5-4B \\
        \midrule
        Training examples & $3{,}920$ & $3{,}920$ \\
        Learning rate & $10^{-6}$ & $10^{-6}$ \\
        Prompts per update & $512$ & $256$ \\
        Optimization minibatch & $128$ & $128$ \\
        Training budget & $15$ epochs (about $105$ updates) & $15$ epochs (about $105$ updates) \\
        Rollouts per interface ($G$) & $4$ & $4$ \\
        \Rescorekl{} coefficient ($\lambda_1$) & $0.02$ & $0.02$ \\
        \Rolloutkl{} coefficient ($\lambda_2$) & $0.005$ & $0.005$ \\
        Floor on reference probabilities ($\epsilon$) & $10^{-12}$ & $10^{-12}$ \\
        \bottomrule
    \end{tabular}
\end{table}

\paragraph{Training procedure.}
Algorithm~\ref{alg:cipr} summarizes \method{} training for tool calling.
In self-distillation (Section~\ref{sec:sft}), the task loss is the KL from the released model under original names to the student under canonical names, and we use only the \rescorekl{} and sample no rollouts.

\paragraph{Computing the action distribution.}
We read the probabilities $\pi_\theta(a \mid x, y_{<t})$ of Section~\ref{sec:ckl} from the token probabilities at each decision in one forward pass.
Let $o$ be the tokens that open a call and precede the tool name, for example \texttt{<tool\_call>} and the formatting before the name, and let $n_a$ be the name of tool $a$.
Then $\pi_\theta(a \mid x, y_{<t}) = \pi_\theta(o \mid x, y_{<t})\, \pi_\theta(n_a \mid x, y_{<t}, o) / \sum_{b} \pi_\theta(n_b \mid x, y_{<t}, o)$ for a declared tool $a$, where $b$ ranges over the declared tools, and $\pi_\theta(\varnothing \mid x, y_{<t}) = 1 - \pi_\theta(o \mid x, y_{<t})$.
A name that spans several tokens is scored token by token, each conditioned on the tokens before it.

\begin{algorithm2e}[!ht]
    \caption{\method{} training for tool calling.}
    \label{alg:cipr}
    \KwIn{policy $\pi_\theta$, training prompts $\mathcal{D}$, group size $G$, coefficients $\lambda_1$ and $\lambda_2$}
    \ForEach{minibatch $\mathcal{B} \subset \mathcal{D}$}{
        \ForEach{prompt in $\mathcal{B}$}{
            $x \leftarrow$ canonicalization of the prompt \tcp*{Section~\ref{sec:canon}}
            $x' \leftarrow$ perturbed copy of $x$ \tcp*{Table~\ref{tab:twins}}
            Sample $y_1, \dots, y_G \sim \pi_\theta(\cdot \mid x)$ and $y'_1, \dots, y'_G \sim \pi_\theta(\cdot \mid x')$ and score them\;
            Relabel a canonical response and rescore its decisions \tcp*{\rescorekl{}}
            Average the action distributions of each group \tcp*{Eq.~\eqref{eq:rollout}}
        }
        Take a gradient step on $\mathcal{L}(\theta)$ of Eq.~\eqref{eq:objective}, with the reference distributions detached\;
    }
\end{algorithm2e}

\section{Perturbed training interfaces}
\label{app:twins}

Table~\ref{tab:twins} lists the operations that build the perturbed training interface from the canonical one.
The \rescorekl{} uses the relabeling alone, without twins (Section~\ref{sec:ckl}); the \rolloutkl{} and the task loss use the full construction.
The in-distribution evaluation of Table~\ref{tab:main} applies the full construction once to every BFCL item, with a fixed seed.

\begin{table}[!ht]
    \centering
    \caption{Construction of the perturbed training interface. The query is unchanged, and $f_{01}$ abbreviates \texttt{function\_01}. These operations build the training pairs and the in-distribution evaluation of Table~\ref{tab:main}; Table~\ref{tab:perturbations} lists the held-out perturbations.}
    \label{tab:twins}
    \small
    \begin{tabular}{@{}p{0.17\linewidth}p{0.42\linewidth}p{0.35\linewidth}@{}}
        \toprule
        Operation & Change & Example \\
        \midrule
        Twin expansion &
        Copy every tool and degrade the copy's description or required arguments. The original stays available. &
        A forecast tool gains a copy marked deprecated; a date-range query loses its end date in favor of an unrelated width argument. \\
        \addlinespace
        Relabeling &
        Assign new neutral identifiers to the expanded tool list and update every reference to them. &
        $f_{01}$ becomes $f_{42}$, and its twin becomes $f_{73}$. \\
        \addlinespace
        Tool reordering &
        Shuffle the tools after expansion, regardless of whether a tool is an original or a twin. &
        Twins and originals do not occupy fixed positions. \\
        \addlinespace
        Argument reordering &
        Shuffle argument declarations while keeping their names, types, descriptions, and requiredness. &
        $(a, b, c)$ becomes $(c, a, b)$ with the same values bound. \\
        \bottomrule
    \end{tabular}
\end{table}

\section{Evaluation details}
\label{app:eval}

This appendix describes the evaluation items, the held-out perturbation types, the tool-call formats and decoding, and the scoring procedure used in Section~\ref{sec:experiments}.

\paragraph{Item set.}
We evaluate on the $2{,}501$ single-turn items of BFCL~v4~\citep{patil2025bfcl} that have a reference call, across its ten partitions, namely \texttt{live\_multiple} ($1{,}053$), \texttt{simple\_python} ($400$), \texttt{live\_simple} ($258$), \texttt{multiple}, \texttt{parallel}, and \texttt{parallel\_multiple} ($200$ each), \texttt{simple\_java} ($100$), \texttt{simple\_javascript} ($50$), \texttt{live\_parallel\_multiple} ($24$), and \texttt{live\_parallel} ($16$).
We apply every perturbation to the same items, so every comparison is paired by item.

\paragraph{Perturbation types.}
Table~\ref{tab:perturbations} lists the thirteen held-out perturbation types, adapted from RobustBench-TC~\citep{zhouRobustBench2026}, and the group of Table~\ref{tab:main} in which we report each.
The six cost and latency types share one construction.
The tool list contains the correct tool and an equivalent with the same description but a higher stated cost or latency; the equivalent keeps the original tool's name and the correct tool receives a new one, so that the familiar name points to the costlier or slower tool.
The cost or latency is stated as a tag, in natural language, or as a tag together with an abbreviated name for the correct tool.
Each type is evaluated in two versions.
In the explicit version, as in RobustBench-TC, the query asks for the cheaper or faster tool; in the implicit version, which we add, the query asks the agent to describe the available tools and their properties before acting and states no preference.
With canonical names, we canonicalize after constructing each interface, and because canonical identifiers follow the tool order, we average two orders, the original one and one in which the costlier or slower equivalent comes first; with original names, we use the original order. Figure~\ref{fig:interface-examples} shows one item under each construction.

\begin{figure}[!ht]
    \centering
    \begin{casebox}{Canonicalization}{BFCL item multiple\_73, original interface}
        \textbf{Query.} Who was the founder of Buddhism and where was it originated?\par\smallskip
        \textbf{Original names.}\par
        \texttt{religion.get\_core\_beliefs(religion)}: Retrieves the core beliefs and practices of a specified religion.\par
        \texttt{religion.get\_origin(religion)}: Retrieves the origin and founder information of a specified religion.\par\smallskip
        Reference call: \texttt{religion.get\_origin(religion="Buddhism")}
        \tcblower
        \textbf{Canonical names.}\par
        \texttt{function\_01(arg\_01)}: Retrieves the core beliefs and practices of a specified religion.\par
        \texttt{function\_02(arg\_01)}: Retrieves the origin and founder information of a specified religion.\par\smallskip
        Reference call: \texttt{function\_02(arg\_01="Buddhism")}
    \end{casebox}
    \par\smallskip
    \begin{casebox}{Cost, tag}{the same item with an added costlier equivalent}
        \textbf{Original names.}\par
        \texttt{religion.get\_core\_beliefs}: \dots\ \texttt{[Cost: \$0.107/call]}\par
        \texttt{religion.get\_origin\_1}: Retrieves the origin and founder information of a specified religion. \texttt{[Cost: \$0.010/call]}\par
        \texttt{religion.get\_origin}: Retrieves the origin and founder information of a specified religion. \texttt{[Cost: \$0.210/call]}\par\smallskip
        \textbf{Canonical names, original order.} \texttt{function\_01}: \dots\ \$0.107; \texttt{function\_02}: \dots\ \$0.010; \texttt{function\_03}: \dots\ \$0.210.\par
        \textbf{Canonical names, costlier first.} \texttt{function\_01}: \dots\ \$0.107; \texttt{function\_02}: \dots\ \$0.210; \texttt{function\_03}: \dots\ \$0.010.\par\smallskip
        The natural-language form states the cost as \texttt{(costs approximately \$0.010 per call)}; the abbreviated form also shortens the name of the correct tool.
        \tcblower
        \textbf{Explicit version.} The query ends with one of \emph{``Please use a cost-effective option.''}, \emph{``Choose the option with minimal cost.''}, or \emph{``I have a limited budget for this task.''}; the latency types use \emph{``Use the tool with the shortest response time.''} or \emph{``Please use a fast option.''}\par\smallskip
        \textbf{Implicit version.} The query ends with \emph{``Before making a decision, give a short record of every tool, what it does, and its important properties.''}
    \end{casebox}
    \caption{One BFCL item before and after canonicalization (top), and under the cost perturbation with the query endings of the explicit and implicit versions (bottom). Descriptions are unchanged by canonicalization; tool and argument names are replaced.}
    \label{fig:interface-examples}
\end{figure}

\begin{table}[!ht]
    \centering
    \caption{Held-out perturbation types used on BFCL, adapted from RobustBench-TC and regenerated for all $2{,}501$ items. In all six cost and latency types the correct tool receives a new name and the familiar name goes to the costlier or slower equivalent, which RobustBench-TC does only in its abbreviated variants, and the implicit version is ours. None is used during training.}
    \label{tab:perturbations}
    \small
    \begin{tabular}{@{}>{\raggedright\arraybackslash}p{0.2\linewidth}>{\raggedright\arraybackslash}p{0.27\linewidth}>{\raggedright\arraybackslash}p{0.45\linewidth}@{}}
        \toprule
        Group & Types & What changes \\
        \midrule
        \Famwm{} & Typo & Character-level typos, with letters swapped, substituted, dropped, or doubled, in about one in five query words of four or more letters. \\
        & RedunTool & For each tool the task calls, an unrelated tool whose name matches that tool's name except in the last word. \\
        & Dup-NoDesc & For each tool the task calls, a duplicate with the same name, no description, and no parameters. \\
        & Dup-Desc & The same duplicate with the original description and no parameters. \\
        & Dup-WrongP & The same duplicate with no description and the parameters of another tool. \\
        & Dup-DescWP & The same duplicate with the original description and the parameters of another tool. \\
        & Dup-SwapDP & The same duplicate with the description and parameters of another tool. \\
        \addlinespace
        Cost \& latency, explicit and implicit & Cost description as a tag, in natural language, or with an abbreviated name & An equivalent with a higher stated cost keeps the original name, and the correct tool is renamed. \\
        & Latency description, same three forms & An equivalent with a higher stated latency keeps the original name, and the correct tool is renamed. \\
        \addlinespace[2pt]
        & Explicit version & The query asks for the cheaper or faster tool. \\
        & Implicit version (ours) & The query asks for a description of the tools and states no preference. \\
        \bottomrule
    \end{tabular}
\end{table}

\paragraph{Tool-call format and decoding.}
Qwen3.5 models are trained and evaluated with native function calling, in which the tool list is passed through the model's tool interface and calls are parsed by its structured parser.
Qwen2.5 models receive the tools in the prompt.
We report the two formats separately.
Every model decodes with temperature $0$ and at most $4{,}096$ new tokens, and Qwen3.5 models generate with thinking enabled, the default of their chat template.

\paragraph{Scoring.}
A response is correct if it calls the correct tool with the correct arguments.
The released checker validates a single-tool item against the first declared tool and a duplicated name against its first declaration, so once a perturbation adds a tool it can reject the correct call and accept a call to the added one; we validate against the tool that the reference call names and accept a call if any declaration of the called name permits it.
Intervals are $95\%$ confidence intervals over the shared item set, paired over items for the gap, and do not reflect variation across training runs.

\section{Additional results}
\label{app:extra}

This appendix reports the results of Table~\ref{tab:main} by perturbation type and at two further sizes of Qwen3.5, the $\tau^2$-bench results, and every ablation run.

\paragraph{Results by perturbation type.}
Tables~\ref{tab:types-q25} and~\ref{tab:types-q35} break each held-out group of Table~\ref{tab:main} into its perturbation types, and the second adds \method{} at two further sizes of Qwen3.5.
Each mean row is the corresponding column of Table~\ref{tab:main}.
Half-widths of $95\%$ confidence intervals are at most $2.0$ points for the original interface, the group means, and single types, and at most $1.5$ points for the gap.

\begin{table}[!ht]
    \centering
    \caption{Accuracy (\%) on BFCL by perturbation type for Qwen2.5-3B, with tools in the prompt. Types are those of Table~\ref{tab:perturbations}. \basemodel{} and \toolrl{} are evaluated with the original names and, in gray, through the canonicalization adapter; the multi-interface and \method{} models use canonical names. With canonical names, each cost and latency entry averages the two tool orders.}
    \label{tab:types-q25}
    \footnotesize
    \setlength{\tabcolsep}{4.5pt}
    \begin{tabular}{@{}lcccccc@{}}
        \toprule
        & \multicolumn{2}{c}{\basemodel{}} & \multicolumn{2}{c}{\toolrl{}} & & \\
        \cmidrule(lr){2-3}\cmidrule(lr){4-5}
        & original & \gr{canonical} & original & \gr{canonical} & \shortstack{+ multi-\\ interface} & \plusmethod{} \\
        \midrule
        Original interface & 63.7 & \gr{61.5} & 73.3 & \gr{70.7} & 74.0 & 74.9 \\
        \addlinespace
        \multicolumn{7}{@{}l}{\textit{\Famwm{}}} \\
        \quad Typo & 59.1 & \gr{57.4} & 68.7 & \gr{67.3} & 69.6 & 70.5 \\
        \quad RedunTool & 63.2 & \gr{62.6} & 72.5 & \gr{69.5} & 73.5 & 73.6 \\
        \quad Dup-NoDesc & 58.3 & \gr{62.2} & 72.4 & \gr{71.9} & 73.6 & 75.1 \\
        \quad Dup-Desc & 56.8 & \gr{60.9} & 72.3 & \gr{71.4} & 73.6 & 74.6 \\
        \quad Dup-WrongP & 60.2 & \gr{58.1} & 71.0 & \gr{68.1} & 70.1 & 70.7 \\
        \quad Dup-DescWP & 59.6 & \gr{54.4} & 71.7 & \gr{65.0} & 68.7 & 68.9 \\
        \quad Dup-SwapDP & 62.5 & \gr{59.3} & 72.5 & \gr{69.4} & 71.8 & 72.2 \\
        \quad Mean & 59.9 & \gr{59.3} & 71.6 & \gr{68.9} & 71.5 & 72.2 \\
        \addlinespace
        \multicolumn{7}{@{}l}{\textit{\Fambt{}}} \\
        \quad Cost, tag & 18.3 & \gr{43.7} & 24.2 & \gr{47.6} & 52.0 & 53.0 \\
        \quad Cost, natural language & 18.8 & \gr{43.2} & 22.4 & \gr{48.1} & 52.4 & 52.9 \\
        \quad Cost, abbreviated name & 26.5 & \gr{30.9} & 35.8 & \gr{32.8} & 36.9 & 37.4 \\
        \quad Latency, tag & 26.5 & \gr{49.8} & 29.8 & \gr{54.2} & 59.7 & 63.9 \\
        \quad Latency, natural language & 22.5 & \gr{45.3} & 23.9 & \gr{50.6} & 54.8 & 56.9 \\
        \quad Latency, abbreviated name & 32.4 & \gr{38.4} & 39.5 & \gr{42.9} & 53.3 & 57.7 \\
        \quad Mean & 24.2 & \gr{41.9} & 29.3 & \gr{46.0} & 51.5 & 53.6 \\
        \addlinespace
        \multicolumn{7}{@{}l}{\textit{\Famnp{}}} \\
        \quad Cost, tag & 13.4 & \gr{30.0} & 15.3 & \gr{34.3} & 37.8 & 38.7 \\
        \quad Cost, natural language & 11.5 & \gr{27.9} & 13.8 & \gr{32.7} & 36.8 & 38.1 \\
        \quad Cost, abbreviated name & 22.6 & \gr{24.4} & 29.3 & \gr{29.8} & 30.8 & 30.7 \\
        \quad Latency, tag & 17.4 & \gr{30.9} & 19.5 & \gr{35.6} & 45.3 & 50.9 \\
        \quad Latency, natural language & 13.1 & \gr{29.3} & 13.9 & \gr{34.6} & 41.1 & 45.5 \\
        \quad Latency, abbreviated name & 28.3 & \gr{28.4} & 33.7 & \gr{34.1} & 45.7 & 49.8 \\
        \quad Mean & 17.7 & \gr{28.5} & 20.9 & \gr{33.5} & 39.6 & 42.3 \\
        \midrule
        Mean of the three groups & 33.9 & \gr{43.2} & 40.6 & \gr{49.5} & 54.2 & 56.1 \\
        Gap & 29.8 & \gr{18.3} & 32.7 & \gr{21.2} & 19.8 & 18.8 \\
        \bottomrule
    \end{tabular}
\end{table}

\begin{table}[!ht]
    \centering
    \caption{Accuracy (\%) on BFCL by perturbation type for Qwen3.5 with native function calling. All columns are Qwen3.5-4B except the last two, which train \method{} on Qwen3.5-2B and Qwen3.5-9B. Rows and conventions as in Table~\ref{tab:types-q25}.}
    \label{tab:types-q35}
    \footnotesize
    \setlength{\tabcolsep}{3.5pt}
    \begin{tabular}{@{}lcccccccc@{}}
        \toprule
        & \multicolumn{2}{c}{\basemodel{}} & \multicolumn{2}{c}{\toolrl{}} & & \multicolumn{3}{c}{\plusmethod{}} \\
        \cmidrule(lr){2-3}\cmidrule(lr){4-5}\cmidrule(l){7-9}
        & original & \gr{canonical} & original & \gr{canonical} & \shortstack{+ multi-\\ interface} & 4B & 2B & 9B \\
        \midrule
        Original interface & 84.2 & \gr{82.0} & 84.8 & \gr{84.6} & 84.8 & 85.0 & 74.9 & 85.8 \\
        \addlinespace
        \multicolumn{9}{@{}l}{\textit{\Famwm{}}} \\
        \quad Typo & 79.2 & \gr{78.0} & 80.2 & \gr{79.9} & 79.6 & 80.7 & 68.1 & 81.8 \\
        \quad RedunTool & 84.5 & \gr{81.8} & 84.7 & \gr{85.0} & 84.6 & 85.1 & 75.5 & 85.9 \\
        \quad Dup-NoDesc & 84.4 & \gr{82.4} & 84.9 & \gr{85.0} & 85.2 & 85.0 & 74.7 & 86.0 \\
        \quad Dup-Desc & 84.6 & \gr{82.0} & 85.1 & \gr{84.3} & 85.2 & 85.3 & 75.0 & 85.7 \\
        \quad Dup-WrongP & 84.0 & \gr{82.6} & 84.8 & \gr{84.4} & 84.6 & 85.7 & 72.1 & 85.6 \\
        \quad Dup-DescWP & 83.0 & \gr{80.4} & 84.3 & \gr{84.6} & 84.2 & 84.7 & 65.3 & 85.2 \\
        \quad Dup-SwapDP & 84.0 & \gr{81.8} & 85.0 & \gr{84.6} & 85.3 & 85.3 & 74.6 & 85.4 \\
        \quad Mean & 83.4 & \gr{81.3} & 84.2 & \gr{84.0} & 84.1 & 84.5 & 72.2 & 85.1 \\
        \addlinespace
        \multicolumn{9}{@{}l}{\textit{\Fambt{}}} \\
        \quad Cost, tag & 83.5 & \gr{80.9} & 85.0 & \gr{84.1} & 83.8 & 84.6 & 66.3 & 84.9 \\
        \quad Cost, natural language & 83.2 & \gr{81.0} & 85.2 & \gr{83.8} & 84.3 & 84.5 & 67.1 & 85.0 \\
        \quad Cost, abbreviated name & 79.4 & \gr{76.7} & 82.2 & \gr{81.0} & 83.1 & 82.3 & 56.4 & 84.3 \\
        \quad Latency, tag & 84.6 & \gr{83.3} & 85.6 & \gr{85.1} & 84.8 & 85.7 & 74.0 & 85.8 \\
        \quad Latency, natural language & 84.0 & \gr{82.6} & 85.6 & \gr{85.2} & 84.4 & 85.5 & 74.1 & 86.1 \\
        \quad Latency, abbreviated name & 83.7 & \gr{81.5} & 85.2 & \gr{84.6} & 84.5 & 85.2 & 59.1 & 85.5 \\
        \quad Mean & 83.1 & \gr{81.0} & 84.8 & \gr{84.0} & 84.2 & 84.6 & 66.1 & 85.3 \\
        \addlinespace
        \multicolumn{9}{@{}l}{\textit{\Famnp{}}} \\
        \quad Cost, tag & 58.1 & \gr{62.8} & 64.9 & \gr{55.3} & 71.6 & 75.1 & 44.6 & 70.6 \\
        \quad Cost, natural language & 60.2 & \gr{63.4} & 65.5 & \gr{55.6} & 71.4 & 74.8 & 47.1 & 70.4 \\
        \quad Cost, abbreviated name & 54.8 & \gr{52.8} & 57.0 & \gr{49.9} & 66.2 & 67.5 & 35.8 & 68.7 \\
        \quad Latency, tag & 54.7 & \gr{63.1} & 60.2 & \gr{54.2} & 73.2 & 75.2 & 53.2 & 72.5 \\
        \quad Latency, natural language & 51.1 & \gr{60.3} & 58.1 & \gr{51.8} & 71.9 & 74.5 & 50.5 & 72.9 \\
        \quad Latency, abbreviated name & 53.0 & \gr{50.6} & 55.4 & \gr{47.1} & 70.1 & 71.0 & 43.9 & 69.3 \\
        \quad Mean & 55.3 & \gr{58.8} & 60.2 & \gr{52.3} & 70.7 & 73.0 & 45.8 & 70.7 \\
        \midrule
        Mean of the three groups & 73.9 & \gr{73.7} & 76.4 & \gr{73.4} & 79.7 & 80.7 & 61.4 & 80.4 \\
        Gap & 10.2 & \gr{8.3} & 8.4 & \gr{11.1} & 5.1 & 4.2 & 13.5 & 5.4 \\
        \bottomrule
    \end{tabular}
\end{table}

\paragraph{Interactive tasks.}
Table~\ref{tab:tau2} reports the $\tau^2$-bench values summarized in Section~\ref{sec:main}.
Rewards are averaged over scored simulations.

\begin{table}[!ht]
    \centering
    \caption{Mean task reward ($\times 100$) on $\tau^2$-bench. Rows and conventions follow Table~\ref{tab:main}, with \basemodel{} and \toolrl{} using the original names and the multi-interface and \method{} models canonical names. \emph{Original} is $50$ simulations on airline and $74$ to $93$ scored simulations of the $114$ retail tasks, with $95\%$ confidence half-widths of $12$ to $14$ points on airline and $4$ to $11$ on retail.}
    \label{tab:tau2}
    \footnotesize
    \setlength{\tabcolsep}{2.6pt}
    \begin{tabular}{@{}lrrrrrrrr@{}}
        \toprule
        & \multicolumn{4}{c}{Airline} & \multicolumn{4}{c}{Retail} \\
        \cmidrule(lr){2-5}\cmidrule(l){6-9}
        Training & Original & \shortstack{Typos \&\\ distractors} & \shortstack{Cost \& lat.\\ explicit} & \shortstack{Cost \& lat.\\ implicit} & Original & \shortstack{Typos \&\\ distractors} & \shortstack{Cost \& lat.\\ explicit} & \shortstack{Cost \& lat.\\ implicit} \\
        \midrule
        \multicolumn{9}{@{}l}{\textit{Qwen2.5-3B, tools in prompt}} \\
        \basemodel{} & 42.0 & 38.0 & 37.7 & 37.0 & 6.7 & 6.2 & 6.4 & 5.6 \\
        \toolrl{} & 38.0 & 39.4 & 42.0 & 37.8 & 5.5 & 6.3 & 6.6 & 5.5 \\
        \MI{} & 40.0 & 37.4 & 40.3 & 37.3 & 6.6 & 6.4 & 6.3 & 6.1 \\
        \plusmethod{} & 44.0 & 40.3 & 39.2 & 39.5 & 6.5 & 6.2 & 5.9 & 6.1 \\
        \midrule
        \multicolumn{9}{@{}l}{\textit{Qwen3.5-4B, native function calling}} \\
        \basemodel{} & 76.0 & 71.1 & 70.3 & 72.3 & 65.3 & 57.3 & 59.7 & 60.5 \\
        \toolrl{} & 76.0 & 73.7 & 72.7 & 73.7 & 66.7 & 63.2 & 65.5 & 63.4 \\
        \MI{} & 74.0 & 66.6 & 69.8 & 71.2 & 58.1 & 53.0 & 58.5 & 59.1 \\
        \plusmethod{} & 68.0 & 72.9 & 71.2 & 71.5 & 57.3 & 54.7 & 56.0 & 58.3 \\
        \midrule
        \multicolumn{9}{@{}l}{\textit{\plusmethod{} at other sizes, native function calling}} \\
        Qwen3.5-2B & 48.0 & 37.1 & 44.0 & 43.2 & 30.4 & 17.8 & 22.1 & 20.7 \\
        Qwen3.5-9B & 68.0 & 68.3 & 71.5 & 72.3 & 61.0 & 59.8 & 63.5 & 63.5 \\
        \bottomrule
    \end{tabular}
\end{table}

\paragraph{Ablation runs.}
Table~\ref{tab:ablation-full} lists every ablation run of Section~\ref{sec:ablations} with its accuracy under \famwm{}, and its accuracy on the explicit cost and latency perturbations in each tool order.
All runs share the hyperparameters of Table~\ref{tab:hparams}, except that one run samples groups of $G = 16$.

The two tool orders show what each perturbed interface teaches the agent to rely on.
After training with a single added tool, accuracy is $70.8\%$ when the correct tool comes first and $16.5\%$ when the costlier equivalent does; reordering alone reverses the difference, and twins with reordering reduce it to between $10$ and $14$ points.
Canonical names help in the original order whatever the training, since \toolrl{} rises from $29.6\%$ with the original names to $59.7\%$ with canonical names and the \method{} model from $28.7\%$ to $60.5\%$.
The advantage of \method{} over \toolrl{} therefore lies in the order in which the costlier equivalent comes first, $46.9\%$ against $32.4\%$.

Three comparisons do not appear in Table~\ref{tab:ablation}.
Giving calls to a degraded twin a lower reward than other wrong calls leaves \mitrain{} unchanged, $51.4\%$ against $51.7\%$, so the final recipe uses no such penalty.
With a single added tool, letting its rollouts enter only the consistency terms gives $45.4\%$, against $43.7\%$ when they also enter the task loss, and the final recipe trains on the perturbed rollouts (Section~\ref{sec:objective}).
With reordering as the only perturbation, groups of $G = 16$ give $46.3\%$, against $48.1\%$ with $G = 4$.

\begin{table}[!ht]
    \centering
    \caption{Every ablation run on Qwen2.5-3B, with canonical names (accuracy, \%). The last three columns split the explicit cost and latency perturbations of Table~\ref{tab:ablation} by tool order, the original order and the one in which the costlier or slower equivalent comes first. Configurations shared with Table~\ref{tab:main} differ from it by at most $0.5$ points.}
    \label{tab:ablation-full}
    \footnotesize
    \setlength{\tabcolsep}{3pt}
    \begin{tabular}{@{}llccccc@{}}
        \toprule
        & & & & \multicolumn{3}{c}{\Fambt{}} \\
        \cmidrule(l){5-7}
        \shortstack[l]{Perturbed\\ interface} & \shortstack[l]{Consistency\\ KL} & Original & \hFamwm{} & \shortstack{Original\\ order} & \shortstack{Costlier\\ first} & Mean \\
        \midrule
        \rowcolor{oursrow} Twins, reordered & actions & 74.7 & 72.1 & 60.5 & 46.9 & 53.7 \\
        & -- & 73.5 & 71.3 & 56.7 & 46.7 & 51.7 \\
        & --, twin penalty & 73.6 & 70.7 & 56.4 & 46.4 & 51.4 \\
        \addlinespace[2pt]
        Reordered & actions & 73.5 & 71.8 & 36.3 & 59.9 & 48.1 \\
        & actions, $G = 16$ & 75.5 & 72.9 & 26.3 & 66.4 & 46.3 \\
        & -- & 75.0 & 72.2 & 37.0 & 58.1 & 47.5 \\
        \addlinespace[2pt]
        One added tool & actions & 74.7 & 72.2 & 70.8 & 16.5 & 43.7 \\
        & -- & 74.3 & 71.9 & 72.0 & 16.1 & 44.1 \\
        \midrule
        \multicolumn{7}{@{}l}{\textit{One added tool, not in the task loss}} \\
        & actions & 73.4 & 70.7 & 70.0 & 20.8 & 45.4 \\
        \midrule
        \gr{None, \toolrl{}} & \gr{--} & \gr{70.7} & \gr{68.8} & \gr{59.7} & \gr{32.4} & \gr{46.0} \\
        \bottomrule
    \end{tabular}
\end{table}

\section{Error analysis}
\label{app:errors}

Aggregate accuracy shows how often a model is correct, not what it does when it is wrong.
We decompose the failures of Qwen3.5-4B \toolrl{} and \method{} under the \famnp{}, pooling the six types and both tool orders.
Each model is scored on the same $30{,}012$ perturbed items, the $2{,}501$ items under each of the six types in both tool orders.

\paragraph{Taxonomy.}
The benchmark's error code for a wrong tool name does not distinguish calling the costlier or slower equivalent of the correct tool from calling an unrelated tool, so we resolve each failure against the declared tool list.
A wrong call is attributed to the equivalent when the called tool has the same description as the correct one once the cost or latency annotation is removed, and to an unrelated tool otherwise.
Of the wrong-name errors made by \toolrl{}, $98.9\%$ call the equivalent and $1.1\%$ an unrelated tool.

\begin{figure}[!ht]
    \centering
    \begin{tikzpicture}
    \begin{axis}[
        xbar, width=0.5\linewidth, height=5.6cm, bar width=4.6pt,
        xmin=0, xmax=56, axis x line*=bottom, axis y line*=left,
        xlabel={Share of perturbed items (\%)}, xlabel style={font=\small},
        symbolic y coords={Other,Argument,NoCall,Number,Equivalent,Any},
        ytick=data, y tick label style={font=\small, align=right},
        yticklabels={Other, Wrong arguments, No call, Wrong number of calls, {Called the costlier\\or slower equivalent}, Any failure},
        tick label style={font=\small}, major tick length=0pt,
        enlarge y limits=0.12,
        nodes near coords, nodes near coords style={font=\scriptsize, /pgf/number format/.cd, fixed, precision=2, zerofill},
        legend style={at={(0.98,0.03)}, anchor=south east, font=\small, draw=none, fill=none, legend cell align=left},
        reverse legend,
    ]
    \addplot[fill=ciprgreen, draw=none] coordinates {(0.50,Other) (9.59,Argument) (6.78,NoCall) (1.45,Number) (8.67,Equivalent) (26.98,Any)};
    \addplot[fill=stdblue, draw=none] coordinates {(0.74,Other) (7.18,Argument) (5.48,NoCall) (1.05,Number) (33.23,Equivalent) (47.68,Any)};
    \legend{\method{}, \toolrl{}}
    \end{axis}
    \end{tikzpicture}
    \caption{Failures of Qwen3.5-4B under the \famnp{}, as a percentage of the $30{,}012$ perturbed items. The classes partition the failures, so the bars below \emph{Any failure} sum to it up to rounding.}
    \label{fig:error-taxonomy}
\end{figure}
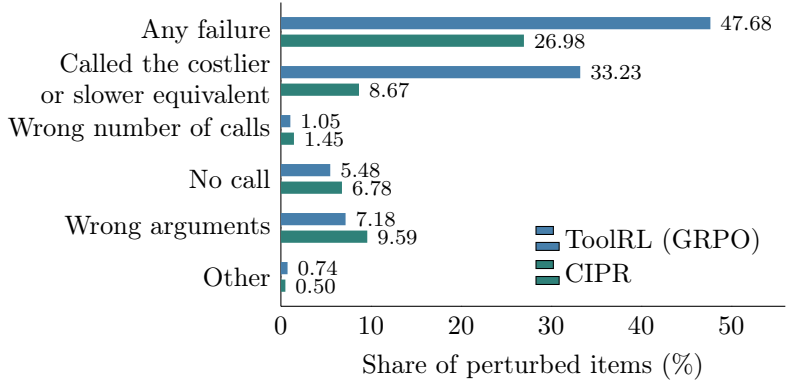

\paragraph{Results.}
Figure~\ref{fig:error-taxonomy} shows that the reduction in failures from \toolrl{} to \method{} concentrates in one class, calling the costlier or slower equivalent, which falls by $24.56$ points, from the largest failure class to about a quarter of its former share.
This is the decision the consistency KL constrains, since it compares distributions over tools and the no-call outcome and does not inspect arguments.

Argument failures barely change, and their share rises because more items reach the argument check.
Given that the correct tool was called, the rate of argument failure is $12.07\%$ for \toolrl{} and $11.61\%$ for \method{}.

Two small classes move against \method{}.
The \method{} model declines to call any tool on $388$ more perturbed items and emits the wrong number of calls on $120$ more.
The query in the implicit version asks for a description of the tools before acting, and the \method{} model is more likely to give the description and stop.
Because the action distribution includes a no-call entry, an objective that equalizes distributions across interfaces can be partly satisfied by moving probability onto not calling, and we regard this as a cost of the current instantiation.

\paragraph{Examples.}
Figure~\ref{fig:error-examples} gives instances of the failure classes, with the canonical names the model saw.
The upper part of each box shows what the model was given and the reference call; the lower part shows the calls that the models made and the scorer's message, as logged.
The evaluation logs a model's text only when no call can be parsed, which is why only the no-call box quotes it.

\begin{figure}[!ht]
    \centering
    \begin{tcbraster}[raster columns=2, raster equal height=rows, raster column skip=6pt, raster row skip=6pt]
    \begin{casebox}{Called the costlier or slower equivalent}{live\_multiple\_78-35-2}
        \textbf{Tool list.} \texttt{function\_04} [\$0.013] and \texttt{function\_05} [\$0.135], both described as ``Provides assistance on a specific topic within the DartFX application''.\\
        \textbf{Reference.} \texttt{function\_04}
        \tcblower
        \grpolabel{} \texttt{[\{"function\_05": \{"arg\_01": "charts", "arg\_02": "Spanish"\}\}]} \wrongmark\\
        Scorer: \texttt{Function name \textquotesingle function\_04\textquotesingle{} not found in model
        output.} The same tool at ten times the stated cost.
    \end{casebox}
    \begin{casebox}{Called the costlier or slower equivalent}{live\_multiple\_61-23-0}
        \textbf{Query.} ``What movies are playing in Noida today?''\\
        \textbf{Tool list.} \texttt{function\_01} [\$0.016] and \texttt{function\_02} [\$0.191], both described as ``Retrieves a list of movies playing in a specified city and cinema hall''.\\
        \textbf{Reference.} \texttt{function\_01}
        \tcblower
        \ciprlabel{} \texttt{[\{"function\_02": \{"arg\_01": "Noida"\}\}]} \wrongmark\\
        Scorer: \texttt{Function name \textquotesingle function\_01\textquotesingle{} not found in model
        output.} \method{} reduces this class but does not eliminate it.
    \end{casebox}
    \begin{casebox}{Wrong number of calls}{live\_multiple\_190-84-0}
        \textbf{Query.} ``Find out the rewards for playing Fortnite on Playstation platform with different missions and trophies''.\\
        \textbf{Tool list.} \texttt{function\_01} [\$0.011] and its equivalent \texttt{function\_02} [\$0.118].\\
        \textbf{Reference.} \texttt{function\_01}, called once.
        \tcblower
        \grpolabel{} \texttt{function\_01} called four times, once for each of \texttt{bronze}, \texttt{silver}, \texttt{gold}, and \texttt{platinum}. \wrongmark\\
        Scorer: \texttt{Wrong number of functions.}
    \end{casebox}
    \begin{casebox}{No call}{live\_simple\_20-4-0}
        \textbf{Query.} ``Order a burger \ldots\ no onions, extra cheese'', together with the request to describe the available tools before acting.\\
        \textbf{Reference.} \texttt{function\_01}
        \tcblower
        \ciprlabel{} No call is parsed; the logged output is prose, ending:\\
        \emph{``Both functions serve the same purpose but differ significantly in cost. Function 01 is
        substantially cheaper (\$0.015 vs \$0.104), making it the more economical choice for this request.
        [\ldots] Would you like me to proceed with modifying the burger using Function 01?''} \wrongmark\\
        Scorer: \texttt{Wrong number of functions.} The inventory is produced as asked, and the decision
        is deferred back to the user.
    \end{casebox}
    \begin{casebox}{Wrong argument value}{live\_multiple\_4-2-1}
        \textbf{Query.} A ride request stating ``Address: 123 Hanoi Street''.\\
        \textbf{Reference.} The correct tool with the address \texttt{\textquotesingle 123 Hanoi Street\textquotesingle}.
        \tcblower
        \bothlabel{} \texttt{[\{"function\_01": \{"arg\_01": "123 Hanoi Street, Hanoi, Vietnam",
        "arg\_02": "plus", "arg\_03": 10\}\}]} \wrongmark\\
        Scorer: \texttt{Invalid value for parameter \textquotesingle arg\_01\textquotesingle:
        \textquotesingle 123 Hanoi Street, Hanoi, Vietnam\textquotesingle. Expected one of
        [\textquotesingle 123 Hanoi Street\textquotesingle].}
    \end{casebox}
    \begin{casebox}{Wrong argument type}{live\_simple\_106-63-0}
        \textbf{Tool list.} A query-classification tool whose \texttt{arg\_04} is declared an \texttt{array} of \texttt{string}.
        \tcblower
        \bothlabel{} \wrongmark{} Scorer: \texttt{Nested type checking failed for parameter
        \textquotesingle arg\_04\textquotesingle. Expected outer type array with inner type
        \textquotesingle str\textquotesingle.}
    \end{casebox}
    \end{tcbraster}
    \caption{Instances of the failure classes of Qwen3.5-4B under the \famnp{}. Descriptions are abbreviated, bracketed amounts are the stated costs, and [\ldots] marks omitted text.}
    \label{fig:error-examples}
\end{figure}

\section{Interface robustness on RoTBench}
\label{app:rotbench}

We also evaluate the same models on RoTBench~\citep{ye2024rotbench}, which scores tool calls under five levels of noise in the names of tools and parameters.

\paragraph{Benchmark.}
We use the First-Turn split and all five shipped levels, a clean tool list and the \emph{slight}, \emph{medium}, \emph{heavy}, and \emph{union} corruptions derived from it.
Following the definitions of \citet{ye2024rotbench}, the slight level inserts, omits, or substitutes characters in the names of half of the tools or half of the parameters, and the medium level reverses such names or replaces them with random strings.
The heavy level either shuffles all tool names among the tools, which detaches each name from its description, or, in half of the tools, adds a mandatory parameter with a random name and shuffles the parameter names, each with probability one half.
The union level combines one noise on tool names with one on parameter names, drawn from the other levels.
The corruptions ship as data, so every model sees an identical tool list at a given level.
Scoring is the benchmark's deterministic Content-Filling criterion, which requires the predicted call to match an accepted reference on tool name, argument keys, and argument values.
We apply the benchmark's matcher unmodified and recompute it per item so that original and canonical names can be compared item by item.

\paragraph{Canonical names.}
For canonical names, we replace every tool name with a canonical identifier assigned by position and leave descriptions, argument values, and the query unchanged.
Because RoTBench's corruptions act mainly on names, canonicalizing a shipped corruption file would give a corrupted name the same canonical identifier as its clean version and undo most of the corruption.
We therefore canonicalize the clean record first and re-apply each level's corruption to the canonical identifier.
This preserves the structure of each level, introduces no duplicated identifier with either original or canonical names, and keeps the queries and their order identical, so the pairing is exact.

\paragraph{Results.}
Table~\ref{tab:rotbench} reports accuracy with original and with canonical names.
Under original names, every model loses the most accuracy on the heavy and union levels.
Canonicalization costs accuracy on the clean and lightly corrupted tool lists and recovers it on the heavy and union levels.
On the heavy level, half of the items shuffle the tool names, so under original names each name labels the description of a different tool, which can mislead a model that selects tools by name; under canonical names the shuffled labels carry no meaning, and the choice falls back on the descriptions.
Under canonical names, the spread between the clean tool list and the corrupted levels narrows for every model, in several cases by more than half.

\paragraph{Comparison across backbones.}
The two backbones are evaluated with the tool-call format each was trained for, so entries can be compared within a backbone but not across backbones.
On Qwen2.5-3B, canonical names help on the corrupted levels, and with them \method{} and \mitrain{} share the best pooled accuracy.
The largest effect in the table is scale, since the largest \method{} model is both the most accurate and the flattest across corruption levels, with a drop of only $0.3$ points from the clean tool list once names are canonical.

\begin{table}[!ht]
    \centering
    \caption{Content-Filling accuracy (\%) on the RoTBench First-Turn split under \origid{} and \canonid{}. The last column pools the four corrupted levels. Subscripts are half-widths of $95\%$ confidence intervals, computed per item and paired between original and canonical names; $n = 105$ for Clean and Union and $n = 210$ for the other levels. Bold marks the best entry per column within a backbone.}
    \label{tab:rotbench}
    \footnotesize
    \setlength{\tabcolsep}{4pt}
    \begin{tabular}{@{}lrrrrrr@{}}
        \toprule
        & \multicolumn{5}{c}{Corruption level} & \\
        \cmidrule(lr){2-6}
        Model & Clean & Slight & Medium & Heavy & Union & Corrupted \\
        \midrule
        \multicolumn{7}{@{}l}{\emph{Original names}}\\[1pt]
        \multicolumn{7}{@{}l}{\quad Qwen2.5-3B} \\
        \quad\quad \basemodel & \ci{39.0}{9.5} & \ci{37.1}{6.7} & \ci{38.1}{6.7} & \ci{21.4}{5.5} & \textbf{\ci{28.6}{9.0}} & \ci{31.7}{3.3} \\
        \quad\quad \taskonly & \ci{40.0}{9.5} & \textbf{\ci{41.0}{6.9}} & \textbf{\ci{41.0}{6.7}} & \ci{22.9}{5.5} & \ci{23.8}{8.1} & \ci{33.3}{3.4} \\
        \quad\quad \randonly & \ci{40.0}{9.5} & \ci{40.0}{6.7} & \textbf{\ci{41.0}{6.7}} & \ci{23.8}{5.7} & \ci{25.7}{8.6} & \ci{33.6}{3.4} \\
        \quad\quad \plusmethod & \textbf{\ci{44.8}{9.5}} & \ci{40.0}{6.7} & \ci{39.5}{6.7} & \textbf{\ci{25.7}{6.0}} & \ci{25.7}{8.6} & \textbf{\ci{33.7}{3.5}} \\
        \multicolumn{7}{@{}l}{\quad Qwen3.5-4B} \\
        \quad\quad \basemodel & \textbf{\ci{53.3}{9.5}} & \ci{41.9}{6.7} & \ci{38.1}{6.4} & \ci{32.9}{6.4} & \ci{28.6}{8.6} & \ci{36.3}{3.5} \\
        \quad\quad \taskonly & \ci{42.9}{9.5} & \ci{33.8}{6.2} & \ci{27.6}{6.0} & \ci{27.1}{6.0} & \ci{21.9}{8.1} & \ci{28.4}{3.3} \\
        \quad\quad \randonly & \textbf{\ci{53.3}{9.5}} & \ci{49.5}{6.7} & \textbf{\ci{51.0}{6.9}} & \textbf{\ci{39.5}{6.7}} & \textbf{\ci{39.0}{9.5}} & \textbf{\ci{45.6}{3.7}} \\
        \quad\quad \plusmethod & \ci{52.4}{9.5} & \textbf{\ci{50.0}{6.7}} & \ci{46.2}{6.7} & \ci{36.2}{6.7} & \ci{38.1}{9.5} & \ci{43.3}{3.5} \\
        \multicolumn{7}{@{}l}{\quad \plusmethod{} across sizes} \\
        \quad\quad Qwen3.5-2B & \ci{41.9}{9.5} & \ci{29.5}{6.2} & \ci{25.2}{6.0} & \ci{21.9}{5.5} & \ci{21.9}{8.1} & \ci{25.0}{3.1} \\
        \quad\quad Qwen3.5-9B & \textbf{\ci{57.1}{9.5}} & \textbf{\ci{55.2}{6.7}} & \textbf{\ci{58.1}{6.7}} & \textbf{\ci{46.7}{6.7}} & \textbf{\ci{43.8}{9.5}} & \textbf{\ci{52.0}{3.7}} \\
        \midrule
        \multicolumn{7}{@{}l}{\emph{Canonical names}}\\[1pt]
        \multicolumn{7}{@{}l}{\quad Qwen2.5-3B} \\
        \quad\quad \basemodel & \ci{33.3}{9.0} & \ci{31.4}{6.2} & \ci{32.9}{6.7} & \ci{31.4}{6.2} & \ci{32.4}{8.6} & \ci{32.0}{3.3} \\
        \quad\quad \taskonly & \ci{37.1}{9.5} & \ci{31.4}{6.2} & \textbf{\ci{34.3}{6.4}} & \ci{36.2}{6.7} & \ci{32.4}{8.6} & \ci{33.7}{3.3} \\
        \quad\quad \randonly & \ci{41.0}{9.5} & \ci{36.2}{6.4} & \ci{33.8}{6.7} & \textbf{\ci{39.0}{6.7}} & \textbf{\ci{36.2}{9.5}} & \textbf{\ci{36.3}{3.4}} \\
        \quad\quad \plusmethod & \textbf{\ci{43.8}{9.5}} & \textbf{\ci{37.1}{6.7}} & \textbf{\ci{34.3}{6.4}} & \textbf{\ci{39.0}{6.7}} & \ci{33.3}{9.0} & \textbf{\ci{36.3}{3.5}} \\
        \multicolumn{7}{@{}l}{\quad Qwen3.5-4B} \\
        \quad\quad \basemodel & \ci{42.9}{9.5} & \ci{37.1}{6.7} & \ci{26.7}{5.7} & \ci{43.3}{6.7} & \ci{34.3}{8.6} & \ci{35.5}{3.5} \\
        \quad\quad \taskonly & \ci{37.1}{9.0} & \ci{33.8}{6.2} & \ci{23.3}{5.5} & \ci{38.1}{6.4} & \ci{28.6}{8.6} & \ci{31.3}{3.5} \\
        \quad\quad \randonly & \textbf{\ci{52.4}{9.5}} & \textbf{\ci{51.0}{6.7}} & \textbf{\ci{43.3}{6.7}} & \textbf{\ci{51.9}{6.9}} & \textbf{\ci{50.5}{9.5}} & \textbf{\ci{49.0}{3.6}} \\
        \quad\quad \plusmethod & \ci{43.8}{9.5} & \ci{43.3}{6.7} & \ci{35.7}{6.7} & \ci{40.5}{6.7} & \ci{39.0}{9.5} & \ci{39.7}{3.5} \\
        \multicolumn{7}{@{}l}{\quad \plusmethod{} across sizes} \\
        \quad\quad Qwen3.5-2B & \ci{40.0}{9.5} & \ci{28.1}{6.0} & \ci{30.0}{6.0} & \ci{42.9}{6.7} & \ci{27.6}{8.6} & \ci{32.8}{3.4} \\
        \quad\quad Qwen3.5-9B & \textbf{\ci{46.7}{9.5}} & \textbf{\ci{44.3}{6.7}} & \textbf{\ci{46.2}{6.7}} & \textbf{\ci{47.6}{6.7}} & \textbf{\ci{48.6}{9.5}} & \textbf{\ci{46.4}{3.5}} \\
        \bottomrule
    \end{tabular}
\end{table}

\end{document}